\documentclass[sigconf]{acmart}

\usepackage{amsmath, bm}
\usepackage{multirow, multicol}
\usepackage{makecell, float, diagbox}
\usepackage[noend]{algpseudocode}

\usepackage{enumitem}
\AtBeginDocument{%
  }

\setcopyright{acmlicensed}
\copyrightyear{2018}
\acmYear{2018}
\acmDOI{XXXXXXX.XXXXXXX}
\acmConference[Conference acronym 'XX]{Make sure to enter the correct
  conference title from your rights confirmation emai}{June 03--05,
  2018}{Woodstock, NY}
\acmISBN{978-1-4503-XXXX-X/18/06}

\begin{document}

\title{PanguIR Technical Report for NTCIR-18 AEOLLM Task}

\author{Lang Mei}
\affiliation{%
    \institution{Huawei Cloud BU}
    \country{China}
}
\email{meilang1@huawei.com}

\author{Chong Chen}
\affiliation{%
    \institution{Huawei Cloud BU}
    \country{China}
}
\email{chenchong55@huawei.com}

\author{Jiaxin Mao}
\affiliation{%
    \institution{Gaoling School of Artificial Intelligence, Renmin University of China}
    \city{Beijing}
    \country{China}
}
\email{maojiaxin@gmail.com}

\renewcommand{\shortauthors}{Trovato et al.}

\begin{abstract}
As large language models (LLMs) gain widespread attention in both academia and industry, it becomes increasingly critical and challenging to effectively evaluate their capabilities. Existing evaluation methods can be broadly categorized into two types: manual evaluation and automatic evaluation. Manual evaluation, while comprehensive, is often costly and resource-intensive. Conversely, automatic evaluation offers greater scalability but is constrained by the limitations of its evaluation criteria (dominated by reference-based answers). To address these challenges, NTCIR-18\footnote{https://research.nii.ac.jp/ntcir/ntcir-18/tasks.html\#AEOLLM} introduced the AEOLLM (Automatic Evaluation of LLMs) task, aiming to encourage reference-free evaluation methods that can overcome the limitations of existing approaches.
In this paper, to enhance the evaluation performance of the AEOLLM task, we propose three key methods to improve the reference-free evaluation: 1) Multi-model Collaboration: Leveraging multiple LLMs to approximate human ratings across various subtasks; 2) Prompt Auto-optimization: Utilizing LLMs to iteratively refine the initial task prompts based on evaluation feedback from training samples; and 3) In-context Learning (ICL) Optimization: Based on the multi-task evaluation feedback, we train a specialized in-context example retrieval model, combined with a semantic relevance retrieval model, to jointly identify the most effective in-context learning examples.
Experiments conducted on the final dataset demonstrate that our approach achieves superior performance on the AEOLLM task.
\end{abstract}



\received{20 February 2007}
\received[revised]{12 March 2009}
\received[accepted]{5 June 2009}

\maketitle

\section{Introduction}
In recent years, the rapid development of Large Language Models (LLMs) and Generative AI has attracted widespread attention from both academia and industry. Models such as the series of Deepseek \cite{liu2024deepseekv2, liu2024deepseekv3, guo2025deepseekr1}, QWen \cite{yang2024qwen2, bai2023qwen}, OpenAI \cite{achiam2023gpt}, with their vast number of parameters (often reaching hundreds of billions or even trillions) and transformer-based pre-training strategies \cite{vaswani2017attention, devlin2019bert}, have demonstrated strong capabilities across multiple natural language processing tasks and zero/few-shot learning scenarios.

The rapid advancement of LLMs has introduced a significant challenge in their development: the need for efficient and effective performance evaluation. A reliable evaluation framework for LLMs is crucial, as it not only facilitates the informed selection of the most suitable models for specific tasks but also provides valuable insights for optimizing their performance.
Existing evaluation methods for LLMs can be broadly categorized into two approaches: manual evaluation and automatic evaluation. Manual evaluation involves human annotators directly assessing the quality of responses generated by LLMs. In contrast, automatic evaluation employs standardized metrics and tools to measure model performance. Compared to manual evaluation, automatic evaluation reduces the reliance on extensive human involvement, thereby lowering costs and improving the objectivity of the assessment process. Consequently, the development of efficient and generalizable automatic evaluation methods has become increasingly important.

However, the existing automatic evaluation methods for LLMs still have the following limitations: they heavily rely on pre-labeled ground truth answers (i.e., reference-based answers) to generate evaluation metrics. In reality, most evaluation tasks do not always have ground truth answers, making reference-based automatic evaluation methods difficult to apply in these scenarios. To encourage the development of reference-free evaluation methods, NTCIR-18 introduced the Automatic Evaluation of LLMs (AEOLLM) task, which includes various subtasks such as dialogue generation, text expansion, summary generation, and non-factoid question answering, to comprehensively test different methods.

In this paper, we propose three key methods to improve the reference-free evaluation: 
\begin{itemize}[leftmargin=1em]
\item \textbf{Multi-model Collaboration}: Different LLMs often exhibit variations in their capabilities across various domains. Relying solely on a single large language model can introduce bias or inadequacies in evaluation results. We combine multiple LLMs to produce evaluation metrics that approximate human ratings across several subtasks.
\item \textbf{Prompt Auto-optimization}: Handcrafted initial prompts may fail to fully prompt the capabilities of large language models in evaluation tasks. We iteratively optimize the prompts based on evaluation feedback from available human-annotated data, starting from the initial prompt instructions, to enhance evaluation performance.
\item \textbf{In-context Learning (ICL) Optimization}: The task performance of LLMs is sensitive to the choice of examples, randomly selected examples may result in performance degradation. To select examples that substantially improve task performance, based on the multi-task evaluation feedback, we train a specialized in-context example retrieval model, combined with a semantic relevance retrieval model, to jointly identify the most effective in-context learning examples.
\end{itemize}

Experiments conducted on the final dataset show that our proposed method achieves superior performance on the AEOLLM task.

\section{Problem Formulation}
The AEOLLM task comprises four subtasks: dialogue generation, text expansion, summary generation and non-factoid question answering. Each evaluation record $D^k_i$ under each task $T_{k}$ includes a question $q^k_i$, an answer $a^k_i$, and a human-annotated quality score $s^k_i$ (ranging from 1 to 5) for the question-answer pair.

Considering the AEOLLM task as a multi-task question-answering quality scoring problem based on LLMs, the goal of our method is to construct the task instruction $P_{T_k}$ for each evaluation record, and make the LLM's generated quality score $LLM\left (P_{T_k} \oplus \left\langle q^k_i, a^k_i \right\rangle  \right )$ of question-answer pair as close as possible to human-annotated score $s^k_i$ across multiple tasks. 
This goal can be measured by several metrics (i.e., Accuracy, Kendall's tau, Spearman correlation coefficient) that evaluate the consistency between the LLM results and the human-annotated results.
This approach tries to narrow the gap between LLM-based solutions and human-based solutions in reference-free evaluation tasks. 

\begin{figure}[!htb]
\centering
\includegraphics[width=0.475\textwidth]{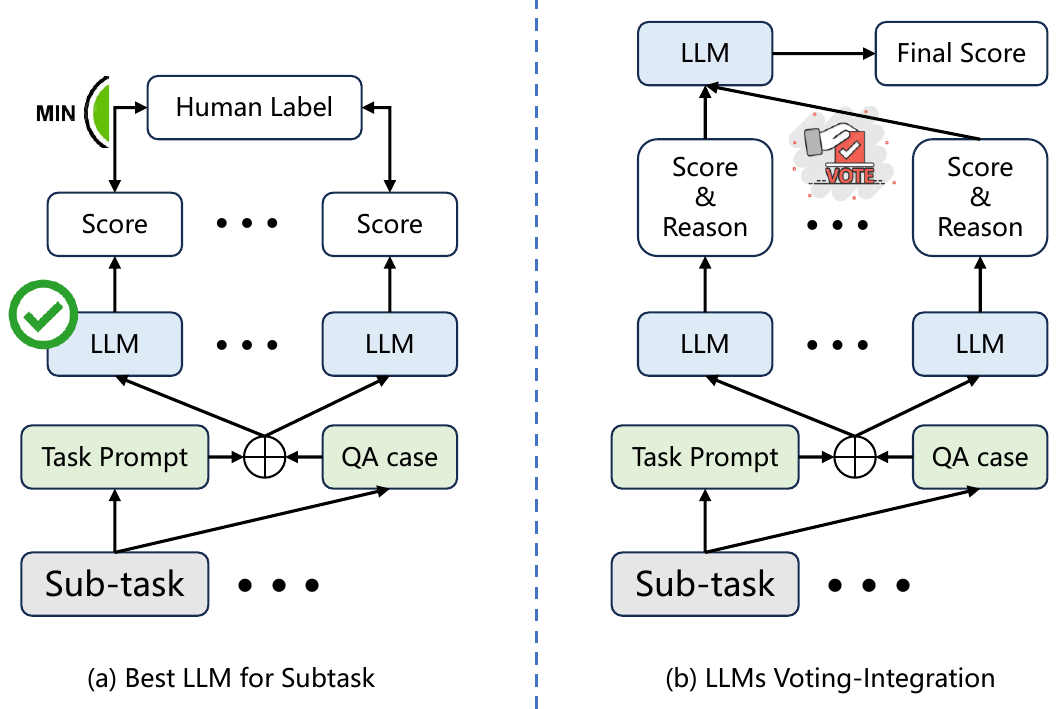}
\caption{In-context Learning (ICL) Optimization for reference-free evaluation.} 
\label{multillm}
\end{figure}

\section{Methodology}
In this section, we provide a detailed introduction to the three proposed methods for improving reference-free evaluation.
\subsection{Multi-model Collaboration}
The performance of LLMs often exhibits significant variability across different domains and tasks, reflecting their diverse architectural designs, training methodologies, and underlying datasets. This heterogeneity in capabilities can lead to systemic biases or performance limitations when evaluations are conducted using a single LLM, as the model's inherent strengths and weaknesses may skew the evaluation results. 

To address these challenges, a more robust and comprehensive approach involves the integration of multiple LLMs to generate aggregated evaluation metrics. By leveraging the complementary strengths of diverse models, this multi-model framework mitigates the risk of individual model biases and provides a more balanced evaluation. The combined metrics derived from this ensemble approach can align more closely with human expert ratings across various subtasks. 

Given a set of LLMs $\left\{LLM_n | n=1,...,N\right\}$, we propose two multi-model collaboration strategies to enhance the performance of reference-free evaluation. In Figure \ref{multillm}, we demonstrate the Multi-model Collaboration for reference-free evaluation.

\subsubsection{Best LLM for Subtask}
Under this strategy, for the evaluation data of each subtask $T_{k}$, we utilize each LLM to generate the quality score $LLM_n\left (P_{T_k} \oplus \left\langle q^k_i, a^k_i \right\rangle \right )$ based on the task instruction $P_{T_k}$, and select the LLM with the best evaluation metrics as the candidate model for executing that subtask.
\begin{equation}
\begin{split}
LLM_{T_k}^{*}=\max_{n} \sum_{i} Metric\left (LLM_n\left (P_{T_k} \oplus \left\langle q^k_i, a^k_i \right\rangle \right ), s^k_i  \right ) 
\end{split}
\end{equation}
\begin{equation}
\begin{split}
\hat{s}_{i}^k = LLM_{T_k}^{*}\left (P_{T_k} \oplus \left\langle q^k_i, a^k_i \right\rangle \right )
\end{split}
\end{equation}

\subsubsection{LLMs Voting-Integration}
Under this strategy, for the evaluation data of each subtask $T_{k}$, we first utilize each LLM to generate the quality score $\hat{s}_{n,i}^k$ and scoring reason $\hat{r}_{n,i}^k$ based on the task instruction $P^k_i$.
Then, a single LLM is used to integrate multiple generated scores and reasons to a final quality score $\hat{s}_{i}^k$.

\begin{figure*}[!htb]
\centering
\includegraphics[width=0.95\textwidth]{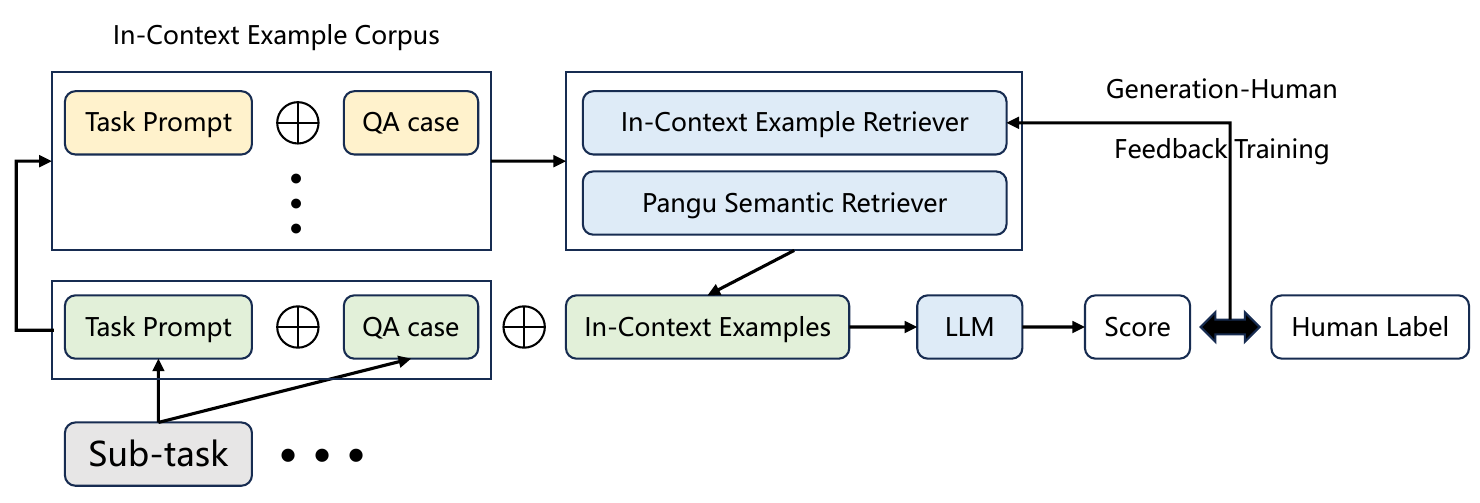}
\caption{Multi-model Collaboration for reference-free evaluation.} 
\label{retriever}
\end{figure*}

\subsection{Prompt Auto-optimization}
Handcrafted initial prompts, typically designed based on prior assumptions or heuristic approaches, often fail to fully eliciting the complete capabilities of LLMs in various evaluation tasks. To address this limitation, we employ an iterative optimization process, leveraging evaluation feedback derived from human-annotated datasets. Guided by evaluation performance, this process begins with the initial prompt instructions, and refines them through multiple iterations.

The iterative optimization framework involves: First, the initial prompt is deployed in the evaluation task to generate the LLM's prediction. Human-annotated data, which serves as a gold standard, provides detailed feedback on the LLM's prediction. This feedback is then analyzed to refine the prompt to better align with the task performance.

Given the initial task prompt $P^{0}_{T_k}$, We iteratively perform the following steps: 1) Sample a certain number (i.e., $M$) of evaluation examples; 2) Generate predicted scores for multiple evaluation examples based on the latest instruction $P^{l}_{T_k}$ and LLM; 3) Use the latest instruction $P^{l}_{T_k}$, multiple evaluation example question-answer pairs, LLM-predicted scores, and human-annotated scores to enable the LLM to generate an improved instruction $P^{l+1}_{T_k}$.

\begin{figure}[!htb]
\centering
\includegraphics[width=0.475\textwidth]{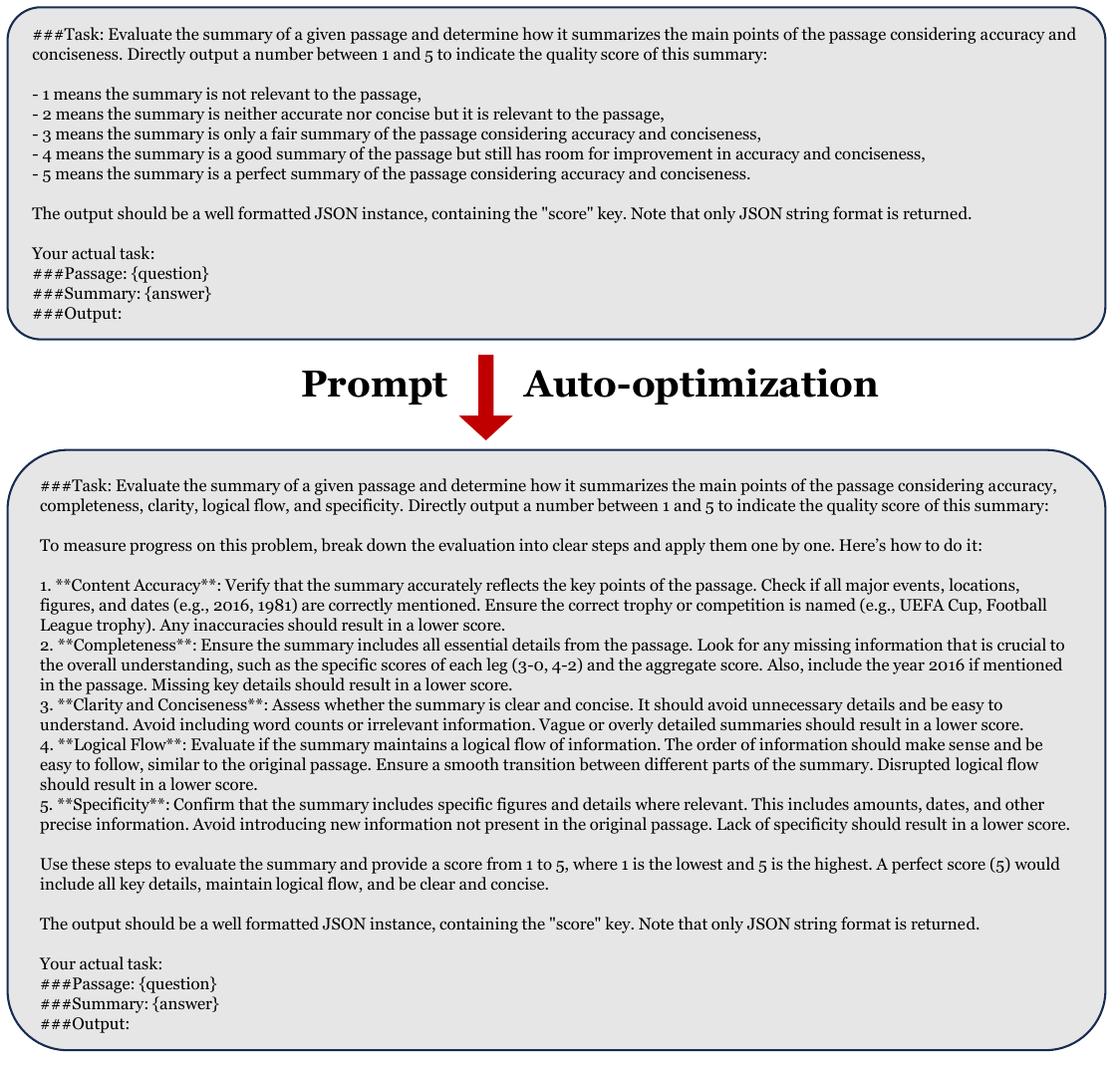}
\caption{The optimized prompt for the summary generation task.} 
\label{optimized_prompt}
\end{figure}

For example, in Figure \ref{optimized_prompt}, we provide the optimized prompt for the summary generation task.

\subsection{In-context Learning (ICL) Optimization}
The performance of LLMs in various tasks is highly sensitive to the selection of in-context examples provided during inference. Empirical evidence suggests that randomly chosen examples can lead to suboptimal performance. To address this challenge, it is critical to develop a systematic approach for identifying high-quality examples that can substantially enhance the model's task performance.

To this end, we propose a novel framework that leverages multi-task evaluation feedback to train a specialized in-context example retrieval model. This model learns to retrieve in-context examples that consistently contribute to better performance across diverse tasks. Furthermore, to ensure the selected examples are not only effective but also semantically relevant to the target task, we integrate a semantic relevance retrieval model into the framework. The combination of these two models (i.e., specialized in-context example retrieval and semantic relevance retrieval) enables a comprehensive and robust approach to identifying the most effective in-context learning examples. In Figure \ref{retriever}, we demonstrate the In-context Learning (ICL) Optimization for reference-free evaluation.

\begin{table*}[htbp]
\centering
\caption{Experimental results of PanguIR during testing, and final phrases.}
\setlength{\tabcolsep}{0.6mm}{
\begin{tabular}{|c|c|c|c|c|c|c|c|c|c|c|c|c|c|c|c|c|} 
    \hline
    \textbf{Phrase} & \textbf{Method} & \multicolumn{3}{|c|}{\textbf{Overall}}  & \multicolumn{3}{|c|}{\textbf{Dialogue}} & \multicolumn{3}{|c|}{\textbf{Text Expansion}} & \multicolumn{3}{|c|}{\textbf{Summary}} & \multicolumn{3}{|c|}{\textbf{Non-factoid QA}} \\
    \hline 
    \multicolumn{2}{|c|}{} & \textbf{Acc.} & \textbf{Ken.} & \textbf{Spear.} & \textbf{Acc.} & \textbf{Ken.} & \textbf{Spear.} & \textbf{Acc.} & \textbf{Ken.} & \textbf{Spear.} & \textbf{Acc.} & \textbf{Ken.} & \textbf{Spear.} & \textbf{Acc.} & \textbf{Ken.} & \textbf{Spear.} \\
    \hline 
    Test & PanguIR & 0.6934 & 0.4618 & 0.5079 & 0.7094 & 0.5272 & 0.5686 & 0.5593 & 0.3680 & 0.4106 & 0.7542 & 0.5039 & 0.5658 & 0.7422 & 0.4481 & 0.4864 \\
    \hline
    \multirow{4}{*}{Final} & PanguIR & \textbf{0.7008} & 0.4579 & 0.4980 & 0.7444 & 0.5611 & 0.6091 & \textbf{0.5581} & 0.3432 & 0.3775 & 0.7479 & 0.5097 & 0.5520 & \textbf{0.7528} & \textbf{0.4175} & \textbf{0.4534} \\ 
    \cline{2-17}
    & UCLWI & 0.6787 & \textbf{0.4704} & \textbf{0.5074} & \textbf{0.7756} & \textbf{0.5798} & \textbf{0.6426} & 0.5266 & 0.3482 & 0.3815 & 0.7273 & \textbf{0.5432} & \textbf{0.5763} & 0.6853 &	0.4105 & 0.4291 \\
    & KNUIR & 0.6654 & 0.4043 &	0.4340 & 0.6778 & 0.4404 & 0.4717 &	0.5512 & 0.3141 & 0.3430 & 0.7375 & 0.4524 & 0.4914 & 0.6951 &	0.4102 & 0.4297 \\
    & ISLab & 0.3225 & 0.2181 &	0.2417 & / & / & / & 0.5241 & \textbf{0.3609} & \textbf{0.4035} & \textbf{0.7658} & 0.5117 & 0.5632 & / & / & /  \\
    \hline
\end{tabular}
}
\label{result}
\end{table*}

\subsubsection{Training Multi-task in-context example retrieval Model}
For each evaluation record $D^k_i$ under each task $T_{k}$, To retrieve the most effective in-context examples, we employed a dual-tower model $E_{icl}$ to encode the current evaluation question-answer pair $\left\langle q^k_i, a^k_i \right\rangle$ and the candidate in-context example $\left\langle q^k_c, a^k_c \right\rangle$, together with the task instruction $P_{T_k}$. Their similarity can be calculated as:
\begin{equation}
\begin{split}
sim\left (\left\langle q^k_i, a^k_i \right\rangle, \left\langle q^k_c, a^k_c \right\rangle \right ) = E_{icl}\left (P_{T_k} \oplus \left\langle q^k_i, a^k_i \right\rangle \right )^TE_{icl}\left (P_{T_k} \oplus \left\langle q^k_c, a^k_c \right\rangle \right )
\end{split}
\end{equation}

To train the in-context example retriever $E_{icl}$,  we introduce a multi-task list-wise ranking training framework based on evaluation feedback. First, we utilize LLM to generate a predicted score augmented with an in-context example, then compute the difference $\left | \hat{s}^k_i - s^k_i \right |$ between the predicted score $\hat{s}^k_i$ and human-annotated score $s^k_i$. A smaller difference indicates a greater contribution of the in-context examples to the evaluation effectiveness. Finally, for each sub-task and each evaluation example, we obtain a ranked list of in-context examples based on their contribution to the evaluation effectiveness.
\begin{equation}
\begin{split}
\hat{s}^k_i = LLM\left (P_{T_k} \oplus \left\langle q^k_i, a^k_i \right\rangle \oplus \left\langle q^k_c, a^k_c \right\rangle \right ) 
\end{split}
\end{equation}

With these in-context examples’ ranks from evaluation feedback, we propose to use the following loss function to inject the ranking signal into the in-context example retriever, inspired by LambdaRank \cite{burges2010ranknet}.
\begin{equation}
\begin{split}
\omega = max\left ( 0, \frac{1}{rank\left ( \left\langle q^k_{c_1}, a^k_{c_1} \right\rangle \right ) } - \frac{1}{rank\left ( \left\langle q^k_{c_2}, a^k_{c_2} \right\rangle \right ) }  \right ) 
\end{split}
\end{equation}
\begin{equation}
\begin{split}
\mathcal{L}=\sum_{k,i,c_1,c_2} \omega * log\left ( 1 + e^{sim\left (\left\langle q^k_i, a^k_i \right\rangle, \left\langle q^k_{c_2}, a^k_{c_2} \right\rangle \right )} - e^{sim\left (\left\langle q^k_i, a^k_i \right\rangle, \left\langle q^k_{c_1}, a^k_{c_1} \right\rangle \right )} \right ) 
\end{split}
\end{equation}

\subsubsection{Semantic Retrieval}
Existing research \cite{liu2021makes} has demonstrated that in-context examples with semantic similarity generally outperform randomly selected examples. Therefore, we adopt semantic retrieval-based examples to enhance the effectiveness of evaluation tasks.
We utilize the semantic retrieval model $E_{Pangu}$ (i.e., Pangu Search) in Huawei Cloud BU to recall the top-ranked in-context examples.

\subsubsection{Diversity integration}
Finally, we combine the top-ranked in-context examples retrieved by in-context example retriever $E_{icl}$ and semantic retrieval model $E_{Pangu}$. In detail, top-ranked in-context examples contain: the most relevant examples with a score $\in$ $\left \{4,5 \right \} $ and $\left \{1,2,3 \right \} $ retrieved by $E_{Pangu}$, the most relevant examples with a score of $\in$ $\left \{4,5 \right \} $ and $\left \{1,2,3 \right \} $ retrieved by $E_{icl}$.

\section{Experiments}
We conduct the experiments on the real-world datasets collected from Didichuxing application, and validate the effectiveness of our proposed framework on the POI re-rank task. 

\subsection{Data Description}
The AEOLLM task comprises four subtasks: dialogue generation, text expansion, summary generation and non-factoid question answering.
Each subtask comprises 700 evaluation data points, with 20\%, 20\%, and 60\% allocated to the training, testing, and final datasets, respectively.
Each evaluation record under each task includes a question, an answer, and a human-annotated quality score (ranging from 1 to 5) for the question-answer pair.

\subsection{Implementation Details}
We adopt the QWen2.5 72B\footnote{https://huggingface.co/Qwen/Qwen2.5-72B-Instruct} and GPT-4o\footnote{https://openai.com/index/hello-gpt-4o/} as our basic LLMs. 
For training the in-context example retrieval model $E_{icl}$, we use gte-large-en-v1.5\footnote{https://huggingface.co/Alibaba-NLP/gte-large-en-v1.5} to initialize the parameters, and use the AdamW \cite{loshchilov2017decoupled} optimizer with a learning rate of 1e-4 to fine-tune $E_{icl}$. 
For each evaluation case, we provide 4 in-context examples to form the input of LLMs.

For evaluation metrics, the AEOLLM task assesses the consistency between the participants' results and the manual annotations using Accuracy, Kendall's tau rank correlation coefficient, and Spearman's rank correlation coefficient as metrics.

\subsection{Experimental Result}
In this section, we demonstrate our experimental results during testing, and final phrases in Table \ref{result}. We observe that our method (PanguIR) achieves the best average performance across the three metrics measuring overall effectiveness, demonstrating the effectiveness of our approach. Notably, PanguIR also attains the best average performance in non-factoid question-answering tasks.



\bibliographystyle{ACM-Reference-Format}
\bibliography{sample-base}

\end{document}